\documentclass[11pt]{article}
\usepackage{acl2016}
\usepackage{times}
\usepackage{url}
\usepackage{latexsym}
\usepackage{xcolor}
\usepackage{balance}
\usepackage{graphicx}
\usepackage{epstopdf}
\usepackage{amsmath}
\usepackage{color}
\usepackage{amsfonts}
\usepackage{amssymb}
\usepackage{booktabs}
\usepackage{enumitem}
\usepackage{mathtools}
\usepackage{multirow}
\usepackage[font=small]{caption}
\usepackage{stmaryrd}

\newcommand{\denselist}{\setlength{\itemsep}{1pt}
  \setlength{\parskip}{0pt} \setlength{\parsep}{0pt}}
\newcommand{\bitem}{\begin{itemize}[noitemsep,topsep=2pt]\denselist}
\newcommand{\eitem}{\end{itemize}}
\newcommand{\qed}{\square}

\newtheorem{theorem}{Theorem}[section]

\DeclareMathOperator*{\argmax}{argmax}

\aclfinalcopy

\title{Globally Normalized Transition-Based Neural Networks}

\author{
  Daniel Andor, Chris Alberti, David Weiss, Aliaksei Severyn, \\
    {\bf Alessandro Presta, Kuzman Ganchev, Slav Petrov and Michael Collins\thanks{$\;\;$On leave from Columbia University.}}\\
  Google Inc\\
  New York, NY\\
  {\footnotesize \tt \{andor,chrisalberti,djweiss,severyn,apresta,kuzman,slav,mjcollins\}@google.com}
}

\date{}

\begin{document}
\maketitle

\begin{abstract}
We introduce a globally normalized transition-based neural network
model that achieves state-of-the-art part-of-speech tagging,
dependency parsing and sentence compression results.  Our model is a
simple feed-forward neural network that operates on a task-specific
transition system, yet achieves comparable or better accuracies than
recurrent models.
We discuss the importance of global as opposed to local normalization:
a key insight is that the label bias problem implies that
globally
normalized models can be strictly more expressive 
than locally normalized models.
\end{abstract}

\graphicspath{{./figs/}}
\newcommand{\bX}{\mathbf{X}}
\newcommand{\bE}{\mathbf{E}}
\newcommand{\bb}{\mathbf{b}}
\newcommand{\bH}{\mathbf{H}}
\newcommand{\bW}{\mathbf{W}}
\newcommand{\bh}{\mathbf{h}}
\newcommand{\mwords}{\mathrm{word}}
\newcommand{\mtags}{\mathrm{tag}}
\newcommand{\mlabels}{\mathrm{label}}
\newcommand{\todo}[1]{{\bf \color{red}{TODO: #1}}}
\newcommand{\eat}[1]{\ignorespaces}
\newcommand{\commentout}[1]{}
\newcommand\T{\rule{0pt}{4ex}}  

\section{Introduction}

Neural network approaches have taken the field of
natural language processing (NLP) by storm.
In particular, variants of long short-term memory (LSTM)
networks \cite{hochreiter1997} have produced impressive
results on some of the classic NLP tasks such as
part-of-speech tagging \cite{ling-EtAl:2015:EMNLP},
syntactic parsing \cite{grammarAsForeign} and
semantic role labeling \cite{zhou-xu:2015:ACL}.
One might speculate that it is the recurrent nature
of these models that enables these results.

In this work we demonstrate that simple feed-forward networks without
any recurrence can achieve comparable or better accuracies than LSTMs,
as long as they are globally normalized.  Our model, described in
detail in Section \ref{sec:model}, uses a transition system
\cite{Nivre:2006} and feature embeddings as introduced by
\newcite{chen-manning:2014:EMNLP}.  We do not use any recurrence, but
perform beam search for maintaining multiple hypotheses and introduce
global normalization with a conditional random field (CRF) objective
\cite{bottou-97,lecun-98h,crf,collobert2011natural}
to overcome the label bias problem that locally normalized models
suffer from.  Since we use beam inference, we approximate
the partition function by summing over the elements in the beam,
and use early updates
\cite{collins-roark:2004:ACL,zhou-etAl:2015:ACL}. We compute gradients
based on this approximate global normalization and perform full
backpropagation training of all neural network parameters based on the
CRF loss.

In Section~\ref{sec:label_bias} we revisit the label bias problem and the implication that
globally normalized models are strictly more expressive than
locally normalized models.
Lookahead features can partially mitigate this discrepancy, but cannot fully
compensate for it---a point to which we return later.
To empirically demonstrate the effectiveness of global normalization, we evaluate our model
on part-of-speech tagging, syntactic dependency parsing
and sentence compression (Section~\ref{sec:experiments}).
Our model achieves state-of-the-art accuracy on all of these tasks,
matching or outperforming LSTMs while being significantly faster.
In particular for dependency parsing on the Wall Street Journal
we achieve the best-ever published unlabeled
attachment score of 94.61\%.

As discussed in more detail in Section \ref{sec:discussion},
we also outperform previous structured training
approaches used for neural network transition-based parsing.
Our ablation experiments show that
we outperform \newcite{weiss-etAl:2015:ACL} and \newcite{alberti-EtAl:2015:EMNLP}
because we do global backpropagation training of all model parameters, while
they fix the neural network parameters when training the global part of their model.
We also outperform \newcite{zhou-etAl:2015:ACL} despite using a smaller beam.
To shed additional light on the label bias problem in practice,
we provide a sentence compression example where the local model completely fails.
We then demonstrate that a globally normalized parsing model without any lookahead
features is almost as accurate as our best model,
while a locally normalized model loses more than 10\% absolute in accuracy
because it cannot effectively incorporate evidence as it becomes available.

Finally, we provide an open-source implementation of our method, called
SyntaxNet,\footnote{http://github.com/tensorflow/models/tree/master/syntaxnet}
which we have integrated into the popular TensorFlow\footnote{http://www.tensorflow.org} 
framework. We also provide a pre-trained,
state-of-the art English dependency parser called ``Parsey McParseface,'' which
we tuned for a balance of speed, simplicity, and accuracy.


\section{Model}
\label{sec:model}
\newcommand{\score}{\rho}
\newcommand{\starts}{s^{\dagger}}
\newcommand{\xt}{{; \theta}}
\newcommand{\xtt}{{\theta}}
\newcommand{\cx}{{}}

At its core, our model is an incremental transition-based parser \cite{Nivre:2006}.
To apply it to different tasks we only need to adjust
the transition system and the input features.

\subsection{Transition System}

Given an input $x$, most often a sentence, we define:
\bitem
\item A set of states $\mathcal S(x)$.
\item A special start state $\starts \in\mathcal S(x)$.
\item A set of allowed decisions ${\cal A}(s, x)$ for all $s\in\mathcal S(x)$.
\item A transition function $t(s, d, x)$ returning a new state $s'$ for
  any decision $d\in {\cal A}(s, x)$.
\eitem 
We will use a function
$\score(s, d, x; \theta)$ to compute the score of decision $d$ in state
$s$ for input $x$. The vector $\theta$ contains the model parameters and we assume
that $\score(s, d, x; \theta)$ is differentiable with respect to
$\theta$.

In this section, for brevity, we will drop the dependence of $x$
in the functions given above, simply writing $\mathcal S$, ${\cal
  A}(s)$, $t(s, d)$, and $\score(s, d; \theta)$.

Throughout this work we will use transition systems in which all
complete structures for the same input $x$ have the same number of
decisions $n(x)$ (or $n$ for brevity).  In dependency parsing for
example, this is true for both the {\em arc-standard} and {\em
  arc-eager} transition systems \cite{Nivre:2006}, where for a
sentence $x$ of length $m$, the number of decisions for any complete
parse is $n(x) = 2 \times m$.\footnote{Note that this is not true for
  the {\em swap} transition system defined in
  \newcite{nivre:2009:ACL}.}  
A complete structure is then a sequence of decision/state pairs $(s_1,
d_1) \ldots (s_n,d_n)$ such that $s_1 = \starts$, $d_i \in {\cal S}(s_i)$ for
$i = 1 \ldots n$, and $s_{i+1} = t(s_i, d_i)$.
We use the notation $d_{1:j}$ to refer to a decision
sequence $d_1 \ldots d_j$.

We assume that there is a
one-to-one mapping between decision sequences $d_{1:j-1}$ and
states $s_j$: that is, we essentially assume that a state
encodes the entire history of decisions.
Thus, each state can be reached by a unique decision
sequence from $\starts$.\footnote{It is straightforward to extend the
  approach to make use of dynamic programming in the case where the
  same state can be reached by multiple decision sequences.}
We will use decision sequences $d_{1:j-1}$ and states
interchangeably: in a slight abuse of notation, we define
$\score(d_{1:j-1}, d; \theta)$ to be equal to $\score(s, d; \theta)$ where $s$ is the
state reached by the decision sequence $d_{1:j-1}$.

The scoring function $\score(s, d; \theta)$ can be defined in a number
of ways. 
In this work, following \newcite{chen-manning:2014:EMNLP}, \newcite{weiss-etAl:2015:ACL},
and \newcite{zhou-etAl:2015:ACL},
we define it via a feed-forward neural network as
\[
\score(s, d; \theta) = \phi(s; \theta^{(l)}) \cdot \theta^{(d)} .
\]
Here $\theta^{(l)}$ are the parameters of the neural network,
excluding the parameters at the final layer. $\theta^{(d)}$ are the
final layer parameters for decision $d$.
$\phi(s; \theta^{(l)})$ is the
representation for state $s$ computed by the neural network under
parameters $\theta^{(l)}$. Note that the score is linear in the
parameters $\theta^{(d)}$. We next describe how
softmax-style normalization can be performed at the local or global level.

\subsection{Global vs.~Local Normalization}

In the \newcite{chen-manning:2014:EMNLP} style of greedy neural
network parsing, the conditional probability distribution over
decisions $d_j$ given context $d_{1:j-1}$ is defined as
\begin{multline}
  \label{eq:local-p}
  p(d_j|d_{1:j-1} \xt) = \frac{\exp \score(d_{1:j-1},d_j \xt) }{Z_L(d_{1:j-1} \xt)} ,
\end{multline}
where
$$
Z_L(d_{1:j-1} \xt) = \sum_{d'\in {\cal A}(d_{1:j-1})} \exp \score(d_{1:j-1},d' \xt)  .
$$
Each $Z_L(d_{1:j-1}; \theta)$ is a {\em local} normalization term.
The probability of a sequence of decisions $d_{1:n}$ is
\begin{align}
  \nonumber
  p_L(d_{1:n} \cx) &= \prod_{j=1}^n p(d_j|d_{1:j-1} \xt) \nonumber \\
  \label{eq:local-beam-p}
  &= \frac{\exp \sum_{j=1}^n \score(d_{1:j-1},d_j \xt)}{
    \prod_{j=1}^n Z_L(d_{1:j-1} \xt)} .
\end{align}
Beam search can be used to attempt to find the maximum of
Eq.~\eqref{eq:local-beam-p} with respect to $d_{1:n}$. The additive scores
used in beam search are the log-softmax of each decision, $\ln
p(d_j|d_{1:j-1} \xt)$, not the raw scores $\score(d_{1:j-1},d_j \xt)$.

In contrast, a Conditional Random Field (CRF) defines a distribution
$p_G(d_{1:n})$ as follows:
\begin{align}
  \label{eq:global-p}
  p_G(d_{1:n} \cx) &= \frac{\exp \sum_{j=1}^n \score(d_{1:j-1},d_j \xt) }{Z_G(\xtt)},
\end{align}
where
\begin{align*}
  Z_G(\xtt) &= \sum_{d'_{1:n}\in {\cal D}_n} \exp \sum_{j=1}^n \score(d'_{1:j-1},d'_j \xt)
\end{align*}
and ${\cal D}_n$ is the set of all valid sequences of decisions of length $n$. 
$Z_G(\xtt)$ is a {\em global} normalization term.
The inference problem is now to find
\begin{align*}
  \argmax_{d_{1:n}\in {\cal D}_n} p_G(d_{1:n} \cx) = 
  \argmax_{d_{1:n}\in {\cal D}_n} \sum_{j=1}^n \score(d_{1:j-1},d_j \xt).
\end{align*}
Beam search can again be used to approximately find the $\argmax$.

\subsection{Training}

Training data consists of inputs $x$ paired with gold decision
sequences $d_{1:n}^*$. We use stochastic gradient descent on the
negative log-likelihood of the data under the model. Under a locally normalized
model, the negative log-likelihood is
\begin{equation}
  \label{eq:local-beam-cost}
\hspace*{-1.4cm}  L_\mathrm{local}(d_{1:n}^*\cx;\theta) = 
  -\ln p_L(d_{1:n}^* \xt) = \end{equation} \\[-1cm]
\[  - \sum_{j=1}^n \score(d_{1:j-1}^*,d_j^* \xt) 
  + \sum_{j=1}^n \ln Z_L(d^*_{1:j-1} \xt) , 
\] 
whereas under a globally normalized model it is
\begin{multline}
  \label{eq:global-cost}
  L_\mathrm{global}(d_{1:n}^*\cx \xt) =
  -\ln p_G(d_{1:n}^* \xt) = \\
  - \sum_{j=1}^n \score(d_{1:j-1}^*,d_j^* \xt) 
  + \ln Z_G(\xtt). \hspace*{1.3cm}
\end{multline}
A significant practical advantange of the locally normalized cost
Eq.~\eqref{eq:local-beam-cost} is that the local partition function $Z_L$ 
and its derivative can usually be computed efficiently.
In contrast, the $Z_G$ term in Eq.~\eqref{eq:global-cost} 
contains a sum over $d'_{1:n}\in {\cal D}_n$ that is in many cases
intractable.

To make learning tractable with the globally normalized model, 
we use beam search and early updates
\cite{collins-roark:2004:ACL,zhou-etAl:2015:ACL}.
As the training sequence is being decoded, we keep track of
the location of the gold path in the beam. If the gold path
falls out of the beam at step $j$, a stochastic gradient
step is taken on the following objective:
\[
\hspace*{-3.0cm}  L_\mathrm{global-beam}(d_{1:j}^*;\theta) = \\
  \]\\[-1cm]
\begin{equation}
\small
\hspace*{-0.2cm}  - \sum_{i=1}^j \score(d_{1:i-1}^*,d_i^* \xt) 
  + \ln \sum_{\mathclap{d'_{1:j}\in {\cal B}_j}}
  \exp \sum_{i=1}^j \score(d_{1:i-1}',d'_i \xt) . \hspace*{-0.2cm}
\label{eq:global-beam}
\end{equation}
Here the set ${\cal B}_j$ contains all paths in the beam at
step $j$, together with the gold path prefix $d^*_{1:j}$.
It is straightforward to derive gradients of the loss 
in Eq.~\eqref{eq:global-beam} and to back-propagate gradients to all
levels of a neural network defining the score $\score(s, d; \theta)$.
If the gold path remains in the beam throughout decoding, a gradient
step is performed using ${\cal B}_n$, the beam at the end of decoding.

\section{The Label Bias Problem}
\label{sec:label_bias}

Intuitively, we would like the model to be able to revise an earlier
decision made during search, when later evidence becomes available
that rules out the earlier decision as incorrect. At first glance, it
might appear that a locally normalized model used in conjunction with
beam search or exact search is able to revise earlier
decisions. However the label bias problem 
(see \newcite{bottou},
\newcite{collins99}
pages 222-226,
\newcite{crf},
\newcite{bottou-lecun-2005}, 
\newcite{smithJohnson07}) means that locally normalized models often have a very
weak ability to revise earlier decisions.

This section gives a formal perspective on the label bias problem,
through a proof that globally normalized models are strictly more
expressive than locally normalized models. The theorem was originally
proved\footnote{More precisely \newcite{smithJohnson07} prove the
  theorem for models with potential functions of the form
  $\score(d_{i-1}, d_i, x_i)$; the generalization to potential
  functions of the form $\score(d_{1:i-1}, d_i, x_{1:i})$ is
  straightforward.}  by \newcite{smithJohnson07}.
The example
underlying the proof gives a clear illustration of the label bias
problem.\footnote{\newcite{smithJohnson07} cite Michael Collins
as the source of the example underlying the proof.  
Note that the theorem refers to {\em conditional} models of the form
$p(d_{1:n} | x_{1:n})$ with global or local normalization.
Equivalence (or non-equivalence) results for {\em joint} models of the
form $p(d_{1:n}, x_{1:n})$ are quite different: for example
results from \newcite{chi99} and \newcite{abneyEtal99} imply that weighted
context-free grammars (a globally normalized joint model) and
probabilistic context-free grammars (a locally normalized joint model)
are equally expressive.}

\paragraph{Global Models can be Strictly More Expressive than
Local Models}
Consider a tagging problem where the task is to map an input sequence
$x_{1:n}$ to a decision sequence $d_{1:n}$.
First, consider a locally normalized model
where we restrict the scoring function to access only the first
$i$ input symbols $x_{1:i}$ when scoring decision $d_i$.
We will return to this restriction soon.
The scoring function $\score$ can be an otherwise arbitrary function of the
tuple $\langle d_{1:i-1}, d_i, x_{1:i} \rangle$:
\begin{align*}
p_L(d_{1:n} | x_{1:n})
&=\prod_{i=1}^n p_L(d_i | d_{1:i-1}, x_{1:i}) \\
&= \frac{\exp \sum_{i=1}^n \score(d_{1:i-1},d_i, x_{1:i}) }
{\prod_{i=1}^n Z_L(d_{1:i-1}, x_{1:i})}.
\end{align*}

Second, consider a globally normalized model
\begin{align*}
&p_G(d_{1:n} | x_{1:n}) 
= \frac{\exp \sum_{i=1}^n \score(d_{1:i-1},d_i, x_{1:i}) }
{Z_G(x_{1:n})}.
\end{align*}
This model again makes use of a scoring function $\score(d_{1:i-1},
d_i, x_{1:i})$ restricted to the first $i$ input symbols when
scoring decision $d_i$.

Define ${\cal P}_L$ to be the set of all possible distributions
$p_L(d_{1:n} | x_{1:n})$ under the local model obtained
as the scores $\score$ vary. Similarly, define ${\cal P}_G$ to be the
set of all possible distributions $p_G(d_{1:n} | x_{1:n})$
under the global model. Here a ``distribution'' is a function
from a pair $(x_{1:n}, d_{1:n})$ to a probability
$p(d_{1:n} | x_{1:n})$.
Our main result is the following:

\begin{theorem} See also \newcite{smithJohnson07}.
\hspace*{0.5cm}${\cal P}_L$ is a strict subset of ${\cal P}_G$, that is ${\cal P}_L
  \subsetneq {\cal P}_G$.
\end{theorem}

To prove this we will first prove that ${\cal P}_L \subseteq {\cal
  P}_G$. This step is straightforward. We then show that
${\cal P}_G \nsubseteq {\cal P}_L$; that is, there are distributions
in ${\cal P}_G$ that are not in ${\cal P}_L$.
The proof that ${\cal P}_G \nsubseteq {\cal P}_L$ gives a clear
illustration of the label bias problem.

{\em Proof that ${\cal P}_L \subseteq {\cal P}_G$:} 
We need to show that for any locally normalized
distribution $p_L$, we can construct a globally normalized model $p_G$
such that $p_G = p_L$.
Consider a locally normalized model with scores
$\score(d_{1:i-1}, d_i, x_{1:i})$.
Define a global model $p_G$ with scores
\[
\score'(d_{1:i-1}, d_i, x_{1:i}) = 
\log p_L(d_i | d_{1:i-1}, x_{1:i}).
\]
Then it is easily verified that 
\[p_G(d_{1:n} | x_{1:n}) = 
p_L(d_{1:n} | x_{1:n}) \]
for all $x_{1:n}, d_{1:n}$.
$\qed$

In proving ${\cal P}_G \nsubseteq {\cal P}_L$ we will use a
simple problem where every example seen in training or test data is
one of the following two tagged sentences:
\begin{align}
x_1 x_2 x_3 = \hbox{a b c}, \;\;d_1 d_2 d_3 = \hbox{A B C}
\nonumber \\
x_1 x_2 x_3 = \hbox{a b e}, \;\;d_1 d_2 d_3 = \hbox{A D E}
\label{eq:lbexample}
\end{align}

Note that the input $x_2 = \hbox{b}$ is ambiguous: it can take tags
$\hbox{B}$ or $\hbox{D}$. This ambiguity is resolved when the next
input symbol, {\tt c} or {\tt e}, is observed.

Now consider a globally normalized model, where the scores
$\score(d_{1:i-1}, d_i, x_{1:i})$ are defined as follows. 
Define ${\cal T}$ as the set $\{ (A, B), (B, C), (A, D), (D, E)\}$
of bigram tag transitions seen in the data. Similarly, define ${\cal
  E}$ as the set $\{ (a, A), (b, B), (c, C), (b, D), (e, E)\}$ of
(word, tag) pairs seen in the data. We define 
\begin{align}
&\score(d_{1:i-1}, d_i, x_{1:i})
\label{eq:alpha} \\
&=\alpha \times \llbracket(d_{i-1}, d_i) \in {\cal T} \rrbracket
+ \alpha \times \llbracket (x_i, d_i) \in {\cal E} \rrbracket
\nonumber
\end{align}
where $\alpha$ is the single scalar parameter of the model,
and $\llbracket \pi \rrbracket = 1$ if $\pi$ is true, $0$ otherwise.

{\em Proof that ${\cal P}_G \nsubseteq {\cal P}_L$:} We
will construct a globally normalized model $p_G$ such that there is
no locally normalized model such that $p_L = p_G$.

Under the definition in Eq.~\eqref{eq:alpha},
it is straightforward to show that
\begin{align*}\small
\lim_{\alpha \rightarrow \infty} p_G(\hbox{A B C} | \hbox{a b c}) =
\lim_{\alpha \rightarrow \infty} p_G(\hbox{A D E} | \hbox{a b e}) = 1 .
\end{align*}

In contrast, under {\em any} definition for $\score(d_{1:i-1}, d_i,
x_{1:i})$, we must have
\begin{equation}
p_L(\hbox{A B C} | \hbox{a b c}) + 
p_L(\hbox{A D E} | \hbox{a b e}) \leq 1
\label{eq:localbad}
\end{equation}
This follows because $p_L(\hbox{A B C} | \hbox{a b c})
= p_L(\hbox{A} | \hbox{a}) \times
p_L(\hbox{B} | \hbox{A}, \hbox{a b}) \times p_L(\hbox{C} | \hbox{A B},
\hbox{a b c})$
and $p_L(\hbox{A D E} | \hbox{a b e})
= p_L(\hbox{A} | \hbox{a}) \times
p_L(\hbox{D} | \hbox{A}, \hbox{a b}) \times
p_L(\hbox{E} | \hbox{A D}, \hbox{a b e})$.
The inequality $p_L(\hbox{B} | \hbox{A}, \hbox{a b}) + p_L(\hbox{D} |
\hbox{A}, \hbox{a b}) \leq 1$ then immediately implies
Eq.~\eqref{eq:localbad}.

It follows that for sufficiently large values of $\alpha$, we have 
$p_G(\hbox{A B C} | \hbox{a b c}) + p_G(\hbox{A D E} | \hbox{a b e}) >
1$, and given Eq.~\eqref{eq:localbad}
it is impossible to define a locally normalized
model with $p_L(\hbox{A B C} | \hbox{a b c}) = p_G(\hbox{A B C} |
\hbox{a b c})$ and $p_L(\hbox{A D E} | \hbox{a b e}) = 
p_G(\hbox{A D E} | \hbox{a b e})$.
$\qed$

\begin{table*}[t]
  \centering%
  \scalebox{1.0}{%
    \small%
    \setlength{\tabcolsep}{4pt}%
    \centering%
    \begin{tabular}{lcccccccccccccccc}
      \toprule
      &\hspace*{0.1cm}& En &\hspace*{0.1cm}& \multicolumn{3}{c}{En-Union} &\hspace*{0.1cm}& \multicolumn{7}{c}{CoNLL '09}&\hspace*{0.1cm}& Avg\\
      Method       && WSJ      &&   News  &   Web   &   QTB   &&    Ca    &   Ch    &   Cz    &   En    &   Ge    &   Ja    &   Sp    &&-  \\
      \midrule
      Linear CRF   &&    97.17 &&   97.60 &   94.58 &   96.04  &&   98.81 &   94.45 &   98.90 &   97.50 &   97.14 &   97.90 &   98.79 &&   97.17\\
      \newcite{ling-EtAl:2015:EMNLP}&& \bf97.78 &&   97.44 &   94.03 &   96.18  &&   98.77 &   94.38 &   99.00 &   97.60 &\bf97.84 &   97.06 &   98.71 &&   97.16\\
      \midrule
      Our Local (B=1) && 97.44 &&   97.66 &   94.46 &   96.59  &&   98.91 &   94.56 &   98.96 &   97.36 &   97.35 &   98.02 &   98.88 &&   97.29\\
      Our Local (B=8) && 97.45 &&   97.69 &   94.46 &   96.64  &&   98.88 &   94.56 &   98.96 &   97.40 &   97.35 &   98.02 &   98.89 &&   97.30\\
      Our Global (B=8)&& 97.44 &&\bf97.77 &\bf94.80 &\bf96.86  &&\bf99.03 &\bf94.72 &\bf99.02 &\bf97.65 &   97.52 &\bf98.37 &\bf98.97 &&\bf97.47\\
            \midrule
      Parsey McParseface\hspace*{-.3cm} && - && 97.52 & 94.24 & 96.45 & - & - & - & - & - & - & - & - && - \\
      \bottomrule
    \end{tabular}
  }
  \caption{\label{tab:pos}
    Final POS tagging test set results on English WSJ and Treebank Union as well as CoNLL'09. We also show the performance of our pre-trained open source model, ``Parsey McParseface.''
}
\end{table*}



Under the restriction that scores $\score(d_{1:i-1}, d_i, x_{1:i})$
depend only on the first $i$ input symbols,
the globally normalized model is still able to model the data in
Eq.~\eqref{eq:lbexample}, while the locally normalized model
fails (see Eq.~\ref{eq:localbad}). The ambiguity at input symbol
{\tt b} is naturally resolved when the next symbol ({\tt c} or {\tt
  e}) is observed, but the locally normalized model is not able to
revise its prediction.

It is easy to fix the locally normalized model for the example
in Eq.~\eqref{eq:lbexample} by
allowing scores $\score(d_{1:i-1}, d_i, x_{1:i+1})$ that 
take into account the input symbol $x_{i+1}$. 
More generally we can have a model of the form $\score(d_{1:i-1}, d_i, x_{1:i+k})$
where the integer $k$ specifies the amount of lookahead in the model.
Such lookahead is common in practice, but insufficient in general.
For every amount of lookahead $k$, 
we can construct examples that cannot be modeled
with a locally normalized model
by duplicating the middle input {\tt b} in (\ref{eq:lbexample}) $k+1$ times.
Only a local model with scores
$\score(d_{1:i-1}, d_i, x_{1:n})$ that considers the entire
input can capture any distribution $p(d_{1:n} | x_{1:n})$:
in this case the decomposition
$ p_L(d_{1:n} | x_{1:n}) = \prod_{i=1}^n p_L(d_i | d_{1:i-1}, x_{1:n}) $
makes no independence assumptions.

However, increasing the amount of context used as input comes
at a cost, requiring more powerful learning algorithms, and
potentially more training data. For a detailed analysis of the trade-offs
between structural features in CRFs and more powerful local classifiers
without structural constraints,
see \newcite{liang08structure}; in these experiments local classifiers
are unable to reach the performance of CRFs on problems such as parsing
and named entity recognition where structural constraints are important.
Note that there is nothing to preclude an approach that makes use of
both global normalization and more powerful scoring functions
$\score(d_{1:i-1}, d_i, x_{1:n})$, obtaining the best of both worlds.
The experiments that follow make use of both.

\section{Experiments}
\label{sec:experiments}

To demonstrate the flexibility and modeling power of our approach, we provide
experimental results on a diverse set of structured prediction tasks.
We apply our approach to POS tagging, syntactic dependency parsing, and sentence
compression.

While directly optimizing the global model defined by Eq.~\eqref{eq:global-cost} works well,
we found that training the model in two steps
achieves the same precision much faster:
we first pretrain the network using the local objective given in Eq.~\eqref{eq:local-beam-cost},
and then perform additional training steps using the global objective given in Eq.~\eqref{eq:global-beam}.
We pretrain all layers except the softmax layer in this way.
We purposefully abstain from complicated hand engineering
of input features, 
which might improve performance further \cite{durrett-klein:2015:ACL}.

We use the training recipe from \newcite{weiss-etAl:2015:ACL} for each training
stage of our model. Specifically, we use averaged stochastic gradient descent
with momentum, and we tune the learning rate, learning rate schedule,
momentum, and early stopping time using a separate held-out corpus for each
task. We tune again with a different set of hyperparameters for training with
the global objective. 

\subsection{Part of Speech Tagging}
\label{subsection:tagging}

Part of speech (POS) tagging is a classic NLP task,
where modeling the structure of the output
is important for achieving state-of-the-art performance.

\paragraph{Data \& Evaluation.}

We conducted experiments on a number of different datasets:
(1) the English Wall Street Journal (WSJ) part
of the Penn Treebank \cite{marcus:1993:CL}
with standard POS tagging splits;
(2) the English ``Treebank Union'' multi-domain corpus containing
data from the OntoNotes corpus version 5 \cite{hovy-EtAl:2006:NAACL},
the English Web Treebank \cite{petrov-mcdonald:2012:SANCL}, and the
updated and corrected Question Treebank \cite{judge-etAl:2006:ACL}
with identical setup to \newcite{weiss-etAl:2015:ACL}; and
(3) the CoNLL '09 multi-lingual shared 
task \cite{hajic-EtAl:2009:CoNLL}.

\paragraph{Model Configuration.}

Inspired by the integrated POS tagging and parsing transition 
system of \newcite{bohnet-nivre:2012:EMNLP-CoNLL}, 
we employ a simple transition system that uses only a {\sc Shift} action and 
predicts the POS tag of the current word on the buffer
as it gets shifted to the stack. 
We extract the following features on a window $\pm 3$ tokens centered
at the current focus token: word, cluster, character n-gram up to length 3.
We also extract the tag predicted for the previous 4 tokens.
The network in these experiments has a single hidden layer with
256 units on WSJ and Treebank Union and 64 on CoNLL'09.

\paragraph{Results.}

In Table \ref{tab:pos} we compare our model
to a linear CRF and to the compositional
character-to-word LSTM model of \newcite{ling-EtAl:2015:EMNLP}.
The CRF is a first-order linear model with exact inference and
the same emission features as our model. It additionally also
has transition features of the
word, cluster and character n-gram up to length 3 on both endpoints of the
transition.
The results for \newcite{ling-EtAl:2015:EMNLP}
were solicited from the authors.

Our local model already compares favorably against these methods on average.
Using beam search with a locally normalized model does not help, but with global normalization
it leads to a 7\% reduction in relative error, empirically demonstrating the effect of label bias.
The set of character ngrams feature is very important, increasing average
accuracy on the CoNLL'09 datasets by about 0.5\% absolute. 
This shows that character-level modeling can also be done with a simple feed-forward
network without recurrence.

\begin{table*}
  \centering
  \scalebox{0.9}{
    \setlength\tabcolsep{5pt}%
    \begin{tabular}{l@{\hskip 0.6cm}ccc@{\hskip 0.7cm}ccc@{\hskip 0.7cm}cc@{\hskip 0.7cm}cc}
      \toprule
      &&\multicolumn{2}{c@{\hskip 0.7cm}}{WSJ} && \multicolumn{2}{c@{\hskip 0.7cm}}{Union-News} & \multicolumn{2}{c@{\hskip 0.9cm}}{Union-Web} & \multicolumn{2}{c@{\hskip 0.0cm}}{Union-QTB\hspace*{0.5cm}}\\
      Method && UAS & LAS && UAS & LAS & UAS & LAS & UAS & LAS\\
      \midrule
      \newcite{martins-etAl:2013:ACL}$^\star$ && 92.89 & 90.55 && 93.10 &  91.13 & 88.23 &  85.04 & 94.21 &  91.54 \\
      \newcite{zhang-mcdonald:2014:ACL}$^\star$ && 93.22 & 91.02 && 93.32 & 91.48 & 88.65 &  85.59 & 93.37 &  90.69 \\
      \newcite{weiss-etAl:2015:ACL} && 93.99 & 92.05 && 93.91 &  92.25 & 89.29 &  86.44 & 94.17 &  92.06 \\
      \newcite{alberti-EtAl:2015:EMNLP} && 94.23 & 92.36 && 94.10 & 92.55 & 89.55 & 86.85 & 94.74 & 93.04 \\
      \midrule
      Our Local (B=1) && 92.95 & 91.02 && 93.11 & 91.46 & 88.42 & 85.58 & 92.49 & 90.38 \\
      Our Local (B=32) && 93.59 & 91.70 && 93.65 & 92.03 & 88.96 & 86.17 & 93.22 & 91.17 \\
      Our Global (B=32) && {\bf 94.61} & {\bf 92.79} && {\bf 94.44} & {\bf 92.93} & {\bf 90.17} & {\bf 87.54} & {\bf 95.40} & {\bf 93.64}  \\
      \midrule
      Parsey McParseface (B=8) && - & - && 94.15 & 92.51 & 89.08 & 86.29 & 94.77 & 93.17 \\
      \bottomrule
    \end{tabular}
  }
  \caption{\label{tab:english_parsing}
    Final English dependency parsing test set results. We note that
    training our system using only the WSJ corpus (i.e. no pre-trained embeddings or other external resources) 
    yields 94.08\% UAS and 92.15\% LAS for our global model with beam 32.
  }
\end{table*}


\begin{table*}[t]
  \centering
  \scalebox{0.9}{%
    \small
    \setlength{\tabcolsep}{2pt}%
    \begin{tabular}{lcc@{\hskip 0.4cm}cc@{\hskip 0.4cm}cc@{\hskip 0.4cm}cc@{\hskip 0.4cm}cc@{\hskip 0.4cm}cc@{\hskip 0.4cm}cc}
      \toprule
      & \multicolumn{2}{c@{\hskip 0.4cm}}{Catalan} & \multicolumn{2}{c@{\hskip 0.4cm}}{Chinese} & \multicolumn{2}{c@{\hskip 0.4cm}}{Czech} & \multicolumn{2}{c@{\hskip 0.4cm}}{English} & \multicolumn{2}{c@{\hskip 0.4cm}}{German} & \multicolumn{2}{c@{\hskip 0.4cm}}{Japanese} & \multicolumn{2}{c@{\hskip 0.4cm}}{Spanish} \\
      Method & UAS & LAS & UAS & LAS & UAS & LAS & UAS & LAS & UAS & LAS & UAS & LAS & UAS & LAS \\
      \midrule
      Best Shared Task Result & - & 87.86 & - & 79.17 & -  & 80.38 & -  & 89.88 & -  & 87.48 & -  & 92.57 & -  & 87.64 \\
      \midrule
      \newcite{BallesterosDS15} & 90.22 & 86.42 & 80.64 & 76.52 & 79.87 & 73.62 & 90.56 & 88.01 & 88.83 & 86.10 & 93.47 & 92.55 & 90.38 & 86.59 \\
      \newcite{zhang-mcdonald:2014:ACL} & 91.41 & 87.91 & 82.87 & 78.57 & 86.62 & 80.59 & 92.69 & 90.01 & 89.88 & 87.38 & 92.82 & 91.87 & 90.82 & 87.34 \\
      \newcite{lei-EtAl:2014:ACL} &  91.33 & 87.22 & 81.67 & 76.71 & 88.76 & 81.77 & 92.75 & 90.00 & 90.81 & 87.81 & {\bf 94.04} & 91.84 & 91.16 & 87.38 \\
  \newcite{bohnet-nivre:2012:EMNLP-CoNLL} & 92.44 & 89.60 & 82.52 & 78.51 & 88.82 & 83.73 &  92.87 & 90.60 & {\bf 91.37} & {\bf 89.38} & 93.67 & 92.63 & 92.24 & 89.60 \\
      \newcite{alberti-EtAl:2015:EMNLP} & 92.31 & 89.17 & 83.57 & 79.90 & 88.45 & 83.57 & 92.70 & 90.56 & 90.58 & 88.20 & 93.99 & {\bf 93.10} & 92.26 & 89.33 \\
      \midrule
      Our Local (B=1) & 91.24 & 88.21 & 81.29 & 77.29 & 85.78 & 80.63 & 91.44 & 89.29 & 89.12 & 86.95 & 93.71 & 92.85 & 91.01 & 88.14 \\
      Our Local (B=16) & 91.91 & 88.93 & 82.22 & 78.26 & 86.25 & 81.28 & 92.16 & 90.05 & 89.53 & 87.4 & 93.61 & 92.74 & 91.64 & 88.88 \\
      Our Global (B=16) & {\bf 92.67} & {\bf 89.83} & {\bf 84.72} & {\bf 80.85} & {\bf 88.94} & {\bf 84.56} & {\bf 93.22} & {\bf 91.23} & 90.91 & 89.15 & 93.65 & 92.84 & {\bf 92.62} & {\bf 89.95} \\
      \bottomrule
    \end{tabular}
  }
  \caption{\label{tab:conll09_final}
    Final CoNLL '09 dependency parsing test set results.}
\end{table*}

\subsection{Dependency Parsing}

In dependency parsing the goal is to produce a directed tree representing
the syntactic structure of the input sentence.

\paragraph{Data \& Evaluation.}

We use the same corpora as in our POS tagging experiments, except that we use
the standard parsing splits of the WSJ. To avoid over-fitting to the development
set (Sec.~22), we use Sec.~24 for tuning the hyperparameters of our models.
We convert the English constituency trees to Stanford style dependencies
\cite{stanford_dependencies} using version 3.3.0 of the converter.
For English, we use predicted POS tags (the same POS tags are used for
all models) and exclude punctuation from the evaluation, as is standard.
For the CoNLL '09 datasets
we follow standard practice and include all punctuation in the evaluation.
We follow \newcite{alberti-EtAl:2015:EMNLP} and 
use our own predicted POS tags so that we can include a k-best tag
feature (see below) but use the supplied predicted morphological features.
We report unlabeled and labeled attachment scores (UAS/LAS).

\paragraph{Model Configuration.}

Our model configuration is basically the same as the one originally proposed
by \newcite{chen-manning:2014:EMNLP} and then refined by
\newcite{weiss-etAl:2015:ACL}. 
In particular, we use the arc-standard
transition system and extract the same set of features
as prior work: words, part of speech tags, and
dependency arcs and labels in the surrounding context of the state, 
as well as k-best tags as proposed by \newcite{alberti-EtAl:2015:EMNLP}.
We use two hidden layers of 1,024 dimensions each.

\paragraph{Results.}

Tables \ref{tab:english_parsing} and \ref{tab:conll09_final} show
our final parsing results and a comparison to the best systems from the literature.
We obtain the best ever published results on almost all datasets, including the WSJ.
Our main results use the same pre-trained word embeddings
as \newcite{weiss-etAl:2015:ACL} and \newcite{alberti-EtAl:2015:EMNLP}, but no tri-training.
When we artificially restrict ourselves to not use pre-trained word embeddings, 
we observe only a modest drop of $\sim$0.5\% UAS;
for example, training only on the WSJ yields 94.08\% UAS and 92.15\% LAS 
for our global model with a beam of size 32.

Even though we do not use tri-training, our model compares favorably to the 94.26\% LAS and 92.41\% UAS
reported by \newcite{weiss-etAl:2015:ACL} with tri-training.
As we show in Sec.~\ref{sec:discussion}, these gains can be attributed to the full backpropagation
training that differentiates our approach from that of \newcite{weiss-etAl:2015:ACL}
and \newcite{alberti-EtAl:2015:EMNLP}.
Our results also significantly outperform the LSTM-based approaches of
\newcite{dyer-etAl:2015:ACL} and \newcite{BallesterosDS15}.



\subsection{Sentence Compression}

Our final structured prediction task is extractive sentence compression.

\paragraph{Data \& Evaluation.}

We follow \newcite{filippova-emnlp15}, where a large news collection is used to
heuristically generate compression instances.
Our final corpus contains about 2.3M compression instances: we use 2M examples
for training, 130k for development and 160k for the final test.
We report per-token F1 score and per-sentence accuracy (A), i.e.~percentage of
instances that fully match the golden compressions.
Following~\newcite{filippova-emnlp15} we also run a human evaluation on 200
sentences where we ask the raters to score compressions for \textit{readability}
(\texttt{read}) and \textit{informativeness} (\texttt{info}) on a scale from 0
to 5.

\paragraph{Model Configuration.}

The transition system for sentence compression is similar to POS
tagging: we scan sentences from left-to-right and label each token as
\textit{keep} or \textit{drop}.
We extract features from words, POS tags, and dependency labels from a window of
tokens centered on the input, as well as features from the history of
predictions.
We use a single hidden layer of size 400.

\begin{table}
  \centering%
  \scalebox{0.85}{%
    \begin{tabular}{lcccc}
      \toprule
      & \multicolumn{2}{c}{Generated corpus} & \multicolumn{2}{c}{Human eval} \\
      Method & A & F1 & read & info \\
      \midrule
      \newcite{filippova-emnlp15} & {\bf 35.36} & {\bf 82.83} & 4.66 & 4.03 \\
      Automatic & - & - & 4.31 & 3.77  \\
      \midrule
      Our Local (B=1) & 30.51 & 78.72 & 4.58 & 4.03 \\
      Our Local (B=8) & 31.19 & 75.69 & - & - \\
      Our Global (B=8) & 35.16 & 81.41 & \textbf{4.67} & \textbf{4.07} \\
      \bottomrule
    \end{tabular}
  }
  \caption{\label{tab:compression-eval}
    Sentence compression results on News data.
    {\em Automatic} refers to application of the same automatic extraction rules used to generate
    the News training corpus.
  }
\end{table}

\paragraph{Results.}

Table~\ref{tab:compression-eval} shows our sentence compression results.
Our globally normalized model again significantly outperforms the local model.
Beam search with a locally normalized model suffers from severe label bias issues
that we discuss on a concrete example in Section~\ref{sec:discussion}.
We also compare to the sentence compression system
from \newcite{filippova-emnlp15}, a 3-layer stacked LSTM which uses dependency
label information.
The LSTM and our global model perform on par on both the automatic evaluation
as well as the human ratings, but our model is roughly 100$\times$ faster.
All compressions kept approximately 42\% of the tokens on average and
all the models are significantly better than the automatic extractions ($p < 0.05$).



\section{Discussion}
\label{sec:discussion}

We derived a 
proof for the label bias problem
and the advantages of global models.
We then emprirically verified this theoretical superiority
by demonstrating state-of-the-art performance on three
different tasks.
In this section we situate and compare our model to
previous work and provide two examples of the label bias problem
in practice.

\subsection{Related Neural CRF Work}

Neural network models have been been combined with 
conditional random fields and globally normalized models before.
\newcite{bottou-97} and \newcite{lecun-98h} describe global training of
neural network models for structured prediction problems.
\newcite{conditional_neural_fields} add a non-linear neural
network layer to a linear-chain CRF and
\newcite{neural_crf} apply a similar approach
to more general Markov network structures.
\newcite{recurrentCRF} and \newcite{Zheng_2015_ICCV}
introduce recurrence into the model and
\newcite{huang2015bidirectional} finally combine
CRFs and LSTMs.
These neural CRF models are limited to
sequence labeling tasks where exact inference is possible,
while our model works well when exact inference is intractable.

\subsection{Related Transition-Based Parsing Work}

For early work on neural-networks for transition-based parsing,
see Henderson \shortcite{henderson:2003:NAACL,henderson:2004:ACL}.
Our work is closest to the work of
\newcite{weiss-etAl:2015:ACL}, \newcite{zhou-etAl:2015:ACL}
and \newcite{watanabe-sumita:2015:ACL};
in these approaches global normalization is added to the
local model of \newcite{chen-manning:2014:EMNLP}.
Empirically, \newcite{weiss-etAl:2015:ACL}
achieves the best performance, even though
their model keeps the parameters of the locally
normalized neural network fixed and only
trains a perceptron that uses the activations as features.
Their model is therefore limited in its ability to
revise the predictions of the locally normalized model.
In Table~\ref{tab:depth} we show that full backpropagation
training all the way to the word embeddings
is very important and significantly contributes
to the performance of our model.
We also compared training under the CRF objective with a 
Perceptron-like hinge loss between the gold and best elements of the beam.
When we limited the backpropagation depth to training only the top layer $\theta^{(d)}$,
we found negligible differences in accuracy:
93.20\% and 93.28\% for the CRF objective and hinge loss respectively.
However, when training with full backpropagation the CRF accuracy 
is 0.2\% higher and training converged more than 4$\times$
faster.

\begin{table}[t]
  \centering
  \scalebox{0.95}{%
    \renewcommand{\arraystretch}{1.0}%
    \setlength\tabcolsep{6pt}%
    \begin{tabular}[h]{lcc}
      \toprule
       Method & UAS & LAS \\
      \midrule
      Local (B=1)         & 92.85 & 90.59 \\
      Local (B=16)        & 93.32 & 91.09 \\
      \midrule
      Global (B=16) $\{\theta^{(d)}\}$    & 93.45 & 91.21 \\
      Global (B=16) $\{W_2, \theta^{(d)}\}$& 94.01 & 91.77 \\
      Global (B=16) $\{W_1, W_2, \theta^{(d)}\}$& 94.09 & 91.81 \\
      Global (B=16) (full)                 & 94.38 & 92.17 \\
      \bottomrule
    \end{tabular}
  }
  \caption{WSJ dev set scores for successively deeper levels of backpropagation.
    The {\em full} parameter set corresponds to backpropagation all the way to the embeddings.
    $W_i$: hidden layer $i$ weights.
  }
  \label{tab:depth}
\end{table}

\begin{table*}[t]
  \centering%
  \small
  \scalebox{0.95}{%
    \setlength{\tabcolsep}{2pt}%
    \begin{tabular}{llcl}
      \toprule
      Method & Predicted compression & $p_L$ & $p_G$ \\
      \midrule
      Local (B=1) & \textcolor{gray}{In Pakistan, former leader} Pervez Musharraf has appeared in court \textcolor{gray}{for the first time, on treason charges}. & $0.13$ & $0.05$\\
      Local (B=8) &\textcolor{gray}{In Pakistan, former leader Pervez Musharraf has appeared in court for the first time, on treason charges}. & {\bf $0.16$} &
$<$$10^{-4}$\\
      Global (B=8) & \textcolor{gray}{In Pakistan, former leader} Pervez Musharraf has appeared \textcolor{gray}{in court for the first time,} on treason charges. & $0.06$  & {\bf $0.07$} \\
      \bottomrule
    \end{tabular}
  }
  \caption{\label{tab:sent-compression-label-bias-example}
    Example sentence compressions where the label bias of the locally normalized 
    model leads to a breakdown during beam search.
    The probability of each compression under the local ($p_L$) and global ($p_G$) models shows that only the global model can properly represent zero probability for the empty compression.
  }
\end{table*}

\newcite{zhou-etAl:2015:ACL} perform full 
backpropagation training like us, 
but even with a much larger beam, their performance is significantly
lower than ours. We also apply our model to two additional tasks,
while they experiment only with dependency parsing.
Finally, \newcite{watanabe-sumita:2015:ACL} introduce recurrent
components and additional techniques like max-violation updates
for a corresponding constituency parsing model.
In contrast, our model does not require any recurrence
or specialized training.

\subsection{Label Bias in Practice}

We observed several instances of severe label bias in the sentence
compression task.  Although using beam search with the local model
outperforms greedy inference on average, beam search leads the local
model to occasionally produce empty compressions
(Table~\ref{tab:sent-compression-label-bias-example}).  It is important to
note that these are {\em not} search errors: the empty compression has
higher probability under $p_L$ than the prediction from greedy
inference. However, the more expressive globally normalized model
does not suffer from this limitation, and correctly gives the empty
compression almost zero probability.

We also present some
empirical evidence that the label bias problem is severe in
parsing. We trained models where the scoring functions in parsing
at position $i$ in the sentence are limited to considering only tokens
$x_{1:i}$; hence unlike the full parsing model, there is no
ability to look ahead in the sentence when making a
decision.\footnote{This setting may be important in some applications,
  where for example parse structures for sentence prefixes are
  required, or where the input is received one word at a time and
  online processing is beneficial.} The result for a greedy model
under this constraint is 76.96\% UAS; for a locally normalized model
with beam search is 81.35\%; and for a globally normalized model
is 93.60\%. Thus the globally normalized model gets very close to the
performance of a model with full lookahead, while the locally
normalized model with a beam gives dramatically lower performance.  In
our final experiments with full lookahead, the globally normalized
model achieves 94.01\% accuracy, compared to 93.07\% accuracy for a
local model with beam search. Thus adding lookahead allows the local
model to close the gap in performance to the global model; however
there is still a significant difference in accuracy, which may in
large part be due to the label bias problem.

A number of authors have considered modified training procedures for
greedy models, or for locally normalized models.
\newcite{daume09searn} introduce Searn, an algorithm that allows a
classifier making greedy decisions to become more robust to errors
made in previous decisions. \newcite{goldberg2013training} describe
improvements to a greedy parsing approach that makes use of methods
from imitation learning \cite{bagnell2011imitation} to augment the
training set. Note that these methods are focused on greedy
models: they are unlikely to solve the label bias problem when used in
conjunction with beam search, given that the problem is one of
expressivity of the underlying model. More recent work
\cite{henderson2015,vaswani2016} has augmented locally normalized
models with {\em correctness probabilities} or {\em error states},
effectively adding a step after every decision where the probability
of correctness of the resulting structure is evaluated. This gives
considerable gains over a locally normalized model, although
performance is lower than our full globally normalized approach.

\section{Conclusions}

We presented a simple and yet powerful model architecture
that produces state-of-the-art results for POS tagging,
dependency parsing and sentence compression.
Our model combines the flexibility of transition-based algorithms and
the modeling power of neural networks.
Our results demonstrate that feed-forward network without
recurrence can outperform recurrent models such as LSTMs
when they are trained with global normalization.
We further support our empirical findings 
with a proof showing that global normalization
helps the model overcome the label bias problem
from which locally normalized models suffer.

\ifaclfinal
\section*{Acknowledgements}

We would like to thank Ling Wang for training his C2W part-of-speech tagger on our setup,
and Emily Pitler, Ryan McDonald, Greg Coppola and
Fernando Pereira for tremendously helpful discussions.
Finally, we are grateful to all members of the Google Parsing Team.
\else\fi

\balance
\bibliographystyle{acl2016}
\bibliography{paper}

\end{document}